\documentclass[letterpaper, 10 pt, journal, twoside]{ieeetran}  

\IEEEoverridecommandlockouts                              

\usepackage{graphics} 
\usepackage{amsmath} 
\usepackage{amssymb}  
\usepackage{algorithm}
\usepackage{algorithmic}
\usepackage{graphicx}
\usepackage{svg}
\usepackage{subfig}
\usepackage{comment}
\usepackage[export]{adjustbox}
\usepackage{hyperref}

\title{
BeTAIL: Behavior Transformer Adversarial Imitation Learning from Human Racing Gameplay}

\author{
Catherine Weaver$^{1}$, Chen Tang$^{1,2}$, Ce Hao$^{1,3}$, Kenta Kawamoto$^4$,  Masayoshi Tomizuka$^{1}$, Wei Zhan$^1$
\thanks{Manuscript received: January 29, 2024; Revised May 1, 2024; Accepted May 31, 2024.}
\thanks{This paper was recommended for publication by Editor Aleksandra Faust upon evaluation of the Associate Editor and Reviewers' comments.} 

\thanks{$^1$Department of Mechanical Engineering, University of California Berkeley, CA, USA. C. Weaver is supported by
NSF GFRP Grant No. DGE 1752814. Contact: \texttt{catherine22@berkeley.edu}}
\thanks{$^2$Department of Computer Science, University of Texas, Austin, USA.}
\thanks{$^3$School of Computing, National University of Singapore, Singapore}
       \thanks{$^4$Sony Research, Tokyo, Japan.}
\thanks{Digital Object Identifier (DOI): see top of this page.}}

\definecolor{myred}{RGB}{184, 84, 80} 
\definecolor{mygreen}{RGB}{106, 145, 83}

\newcommand{\blue}[1]{\textcolor{black}{#1}}
\newcommand{\new}[1]{\textcolor{black}{#1}}

\begin{document}
\bstctlcite{IEEEexample:BSTcontrol}
\markboth{IEEE Robotics and Automation Letters. Preprint Version. Accepted June, 2024}
{Weaver \MakeLowercase{\textit{et al.}}: BeTAIL: Behavior Transformer Adversarial Imitation Learning from Human Racing Gameplay} 
\maketitle

\begin{abstract}
\blue{
Autonomous racing poses a significant challenge for control, requiring planning minimum-time trajectories under uncertain dynamics and controlling vehicles at their handling limits. Current methods requiring hand-designed physical models or reward functions specific to each car or track.
In contrast, imitation learning uses only expert demonstrations to learn a control policy. }Imitated policies must model complex environment dynamics and human decision-making. Sequence modeling is highly effective in capturing intricate patterns of motion sequences but struggles to adapt to new environments or distribution shifts that are common in real-world robotics tasks. In contrast, Adversarial Imitation Learning (AIL) can mitigate this effect, but struggles with sample inefficiency and handling complex motion patterns.
Thus, we propose BeTAIL: Behavior Transformer Adversarial Imitation Learning, 
which combines a Behavior Transformer (BeT) policy from human demonstrations with online AIL. 
BeTAIL adds an AIL residual policy to the BeT policy to model the sequential decision-making process of human experts and correct for out-of-distribution states or shifts in environment dynamics. We test BeTAIL on three challenges with expert-level demonstrations of real human gameplay in \blue{the high-fidelity racing game }Gran Turismo Sport. 
Our proposed BeTAIL reduces environment interactions and improves racing performance and stability, even when the BeT is pretrained on different tracks than downstream learning. Videos and code available at: 
{\footnotesize \url{https://sites.google.com/berkeley.edu/BeTAIL/home}}.

\end{abstract}

\begin{IEEEkeywords}
Imitation Learning, Reinforcement Learning, Deep Learning Methods
\end{IEEEkeywords}


\section{Introduction}
\IEEEPARstart{A}{utonomous} racing is of growing interest to inform controller design at the limits of vehicle handling and provide a safe alternative to racing with human drivers~\blue{\cite{betz2022autonomous}}. An autonomous racer's driving style should resemble \textit{human-like racing behavior} in order to behave in safe and predictable ways that align with the written and social rules of the race \cite{wurman2022outracing}. \blue{ High-fidelity racing simulators, such as the world-leading Gran Turismo Sport (GTS) game, can test policies in a safe and realistic environment and  benchmark comparisons between autonomous systems and human drivers }\cite{wurman2022outracing, fuchs2021super}. Reinforcement learning (RL) outperforms expert human players but requires iterative tuning of dense rewards~\cite{wurman2022outracing}\blue{, which is susceptible to} ad hoc trial and error~\cite{Booth_Knox_Shah_Niekum_Stone_Allievi_2023}.
Imitation learning (IL) is a potential solution that mimics experts' behavior with offline demonstrations \cite{zare2023survey}. \blue{We propose a novel IL algorithm to model non-Markovian decision-making of human experts in GTS.}

Human racing includes complex decision-making and understanding of environment dynamics  \cite{zhu2023gaussian}, and the performance of Markovian policies can deteriorate with human demonstrations \cite{orsini2021matters}. Sequence-based transformer architectures~\cite{chen2021decision, janner2021offline}, similar to language models \cite{radford2018improving}, accurately model the complex dynamics of human thought \cite{li2023guided}. The Behavior Transformer (BeT) \cite{shafiullah2022behavior}, and Trajectory Transformer \cite{janner2021offline}, which do not require pre-defined environment rewards, are casually conditioned on the past to accurately model long-term dynamics \cite{janner2021offline}.  Policies are trained via supervised learning to autoregressively maximize the likelihood of trajectories in the offline dataset. Policies are limited by dataset quality \cite{zheng2022online} and are sensitive to variation in \new{system dynamics and out-of-distribution states.}

Adversarial Imitation Learning (AIL) \cite{ho2016generative} overcomes the issues with offline learning with adversarial training and reinforcement learning.  A discriminator network encourages the agent to match the state occupancy of online rollouts and expert trajectories, reducing susceptibility to distribution shift when encountering unseen states \cite{zare2023survey}. However, AIL requires extensive environment interactions, and its performance  deteriorates with human demonstrations \cite{orsini2021matters}. Thus, AIL in racing is unstable and sample inefficient. AIL also exhibits shaky steering behavior and often spins off the track, since AIL does not model humans' non-Markovian decision-making.

We propose Behavior Transformer Adversarial Imitation Learning (BeTAIL), which leverages offline sequential modeling and online occupancy-matching fine-tuning to 1.) capture the sequential decision-making process of human demonstrators and 2.) correct for out-of-distribution states or minor shifts in environment dynamics. First, a BeT policy is learned from offline human demonstrations. 
Then, an AIL mechanism finetunes the policy to match the state occupancy of the demonstrations. BeTAIL adds a residual policy, e.g. \cite{silver2018residual}, to the BeT action prediction; the residual policy refines the agent's actions while remaining near the action predicted by the BeT. 
Our contributions are as follows:
\begin{enumerate}
\item We propose Behavior Transformer Adversarial Imitation learning (BeTAIL) to pre-train a BeT and fine-tune it with a residual AIL policy to learn complex, non-Markovian behavior from human demonstrations.
\item We show that when learning a racing policy from real human gameplay in Gran Turismo Sport,  BeTAIL outperforms BeT or AIL alone while closely matching non-Markovian patterns in human demonstrations.
\item We show BeTAIL when pre-training on a library of demonstrations from multiple tracks to improve sample efficiency and performance when fine-tuning on an unseen track with a single demonstration trajectory. 
\end{enumerate}
In the following, we discuss related works (Section \ref{sec:related_works}) and preliminaries (Section \ref{sec:prelim}). Then we introduce BeTAIL in Section \ref{sec:method} and describe our method for imitation of human gameplay in Section \ref{sec:experiment_setup}. Finally, Section \ref{sec:results} describes three challenges in GTS with concluding remarks in Section \ref{sec:conclude}.

\section{Related Works}\label{sec:related_works}
\subsection{Behavior Modeling}
Behavior modeling aims to capture human behavior, which is important for robots and vehicles that operate in proximity to humans \cite{brown2020taxonomy}. AIL overcomes the problem of cascading errors with traditional techniques like behavioral cloning and parametric models \cite{kuefler2017imitating}. Latent variable spaces allow researchers to model multiple distinct driving behaviors \cite{li2017infogail, bhattacharyya2022modeling, fernando2018learning, sharma2018directed}. However, sample efficiency and training stability are common problems with AIL from scratch, which is exacerbated when using human demonstrations \cite{orsini2021matters}.
Augmented rewards \cite{li2017infogail, fernando2018learning} or negative demonstrations \cite{lee2020mixgail} can accelerate training but share the same pitfalls as reward shaping in RL.  

\subsection{Curriculum Learning and Guided Learning}
\new{Structured training regimens can accelerate RL.}
 Curriculum learning gradually increases the difficulty of tasks~\cite{song2021autonomous} and can be automated with task phasing, which gradually transitions a dense imitation-based reward to a sparse environment reward \cite{bajaj2023task}. 
 ``Teacher policies'' can accelerate RL through policy intervention to prevents unsafe actions \cite{xue2022guarded, liu2023guide} or guided policy search \cite{pmlr-v28-levine13} to guide the objective of the agent. Thus, large offline policies can be distilled into lightweight RL policies \cite{li2023guided}. BeTAIL uses residual RL policies \cite{zhang2022residual, silver2018residual, johannink2019residual}, which adapt to \new{task variations} by adding a helper policy that can be restricted close to the teacher's action~\cite{rana2023residual,won2022physics}.

\subsection{Sequence Modeling}
Sequence-based modeling in offline RL predicts the next action in a trajectory sequence, which contains states, actions, and optionally goals. Commonly, goals are set to the return-to-go, i.e. the sum of future rewards \cite{chen2021decision, janner2021offline}, but advantage conditioning improves performance in stochastic environments \cite{gao2023act}. Goal-conditioned policies are fine-tuned with online trajectories and automatic goal labeling \cite{zheng2022online}. Sequence models can be distilled into lightweight policies with offline RL and environment rewards~\cite{li2023guided}. 
However, when rewards are not available, e.g. \new{IL} \cite{shafiullah2022behavior, janner2021offline}, offline sequence models suffer from distribution shift and poor dataset quality~\cite{zheng2022online}.

\color{black}
\section{Preliminaries}\label{sec:prelim}
\subsection{Problem Statement}

We model the learning task as a Markov Process (MP) defined by $\{\mathcal{S}, \mathcal{A}, T\}$ of states $s\in\mathcal{S}$, actions $a \in \mathcal{A}$, and the transition probability $T(s_t,a_t,s_{t+1}): \mathcal{S} \times \mathcal{A} \times \mathcal{S} \mapsto[0,1]$. Note that, unlike RL, we do not have access to an environment reward.  We have access to human expert demonstrations in the training environment consisting of a set of trajectories, ${D_E}=(\tau_0^{E}, \tau_1^{E} ..., \tau_M^{E})$, of states and actions at every time step $\tau=(s_t, a_t, ...)$. The underlying expert policy, $\pi_E$, is unknown. The goal is to learn the agent's policy $\pi$, that best approximates the expert policy $\pi_E$.
Each expert trajectory $\tau^E$, consists of states and action pairs: $\tau^E = (s_0, a_0, s_1, a_1, \hdots, s_N, a_N)$.
The human decision-making process of the expert is unknown and likely non-Markovian \cite{mandlekar2021matters}; thus, imitation learning performance can deteriorate with human trajectories~\cite{orsini2021matters}.

\subsection{Unimodal Decision Transformer}
The Behavior Transformer (BeT) processes the trajectory $\tau_E$ as a sequence of 2 types of inputs: states and actions. 
The original BeT implementation \cite{shafiullah2022behavior} employed a mixture of Gaussians to model a dataset with multimodal behavior. For simplicity and to reduce the computational burden, we instead use an unimodal BeT that uses a deterministic similar to the one originally used by the Decision Transformer \cite{chen2021decision}. However, since residual policies can be added to black-box policies \cite{silver2018residual}, BeTAIL's residual policy could be easily added to the k-modes present in the original BeT implementation. 

At timestep $t$, the BeT uses the tokens from the last $K$ timesteps to generate the action $a_t$, where $K$ is referred to as the context length. Notably, the context length during evaluation can be shorter than the one used for training. The BeT learns a deterministic policy $\pi_{\mathrm{BeT}}(a_t|\mathbf{s}_{-K,t})$, where $\mathbf{s}_{-K,t}$ represents the sequence of $K$ past states $\mathbf{s}_{\max (1, t-K+1): t}$. The policy is parameterized using the minGPT architecture \cite{brown2020language},  which applies a causal mask to enforce the autoregressive structure in the
predicted action sequence.

A notable strength of BeT is that the policy can model non-Markovian behavior; in other words, rather than modeling the action probability as $P(a_t|s_t)$, the policy models the probability $P(a_t|s_t, s_{t-1}, ..., s_{t-h+1})$. However, the 
policy is trained using \textit{only} the offline dataset, and even minor differences between the training data and evaluation environment can lead to large deviations in the policies' performance \cite{zare2023survey}.

\section{Behavior Transformer-Assisted Adversarial Imitation Learning}\label{sec:method}
\begin{figure}[t]
    \centering
    \vspace{3pt}
    \includegraphics[width=.35\textwidth]{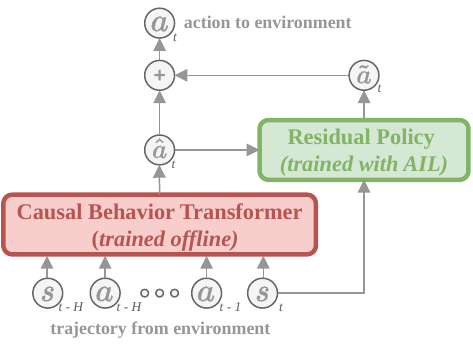}
    \caption{BeTAIL rollout collection. The pre-trained BeT predicts action $\hat{a}_t$ from the last $H$ state-actions. 
    Then the residual policy specifies action $\tilde{a}_t$ from the current state and  $\hat{a}_t$, and the agent executes $a_t=\hat{a}_t+\tilde{a}_t$ in the environment.}
    \label{fig:diagram}
\end{figure}

We now present Behavior Transformer-Assisted Adversarial Imitation Learning (BeTAIL), summarized in Fig. \ref{fig:diagram}. \blue{BeTAIL consists of a pretrained BeT policy and a corrective residual policy \cite{rana2023residual}. First, an unimodal, causal Behavior Transformer (BeT) \cite{shafiullah2022behavior} is trained on offline demonstrations to capture the sequential decision-making of the human experts. Then, the BeT policy is frozen, and a lightweight, residual policy is trained with  Adversarial Imitation Learning (AIL).} Thus, BeTAIL combines offline sequential modeling and online occupancy-matching fine-tuning to capture the decision-making process of human demonstrators and adjust for out-of-distribution states or environment changes. 


\subsection{Behavior Transformer (BeT) Pretraining}\label{sec:pretrain}
A unimodal BeT policy $\hat{a}=
\pi_{\mathrm{BeT}}(a_t|\textbf{s}_{-K,t})$ is used to predict a base action $\hat{a}$. 
Following \cite{chen2021decision, zheng2022online}, sub-trajectories of length $K$ in the demonstrations are sampled uniformly and trained to minimize the difference between the action predicted by the BeT and the next action in the demonstration sequence. 
While the action predicted by the BeT, $\hat{a}$, considers the previous state history $\textbf{s}_{-K,t}$, the action may not be ideal in stochastic environments \cite{gao2023act}. 
\blue{Existing methods to update Transformer models  require rewards and rely on supervised learning or on-policy algorithms \cite{li2023survey}. However, the complex racing tasks requires off-policy RL to converge to the maximum reward \cite{fuchs2021super}. Further, we assume an environment reward is not available; rather, we wish to replicate human demonstrations.  Thus, we freeze the weights of the BeT after the pretraining stage, and add a corrective residual policy \cite{silver2018residual} that can be readily trained with off-policy AIL. }

\subsection{Residual Policy Learning for Online Fine-tuning}\label{sec:betail(r)}
 The agent's action is the sum of the action specified by the BeT and the action from a residual policy \cite{silver2018residual}, which corrects or improves the actions from the BeT base policy. 
We define an augmented MP: $\tilde{\mathcal{M}}=\{\tilde{\mathcal{S}}, \tilde{\mathcal{A}}, \tilde{T}\}$. The state space $\tilde{\mathcal{S}}$ is augmented with the base action: $\tilde{s}_t \doteq \begin{bmatrix}s_t & a_t
        \end{bmatrix}^\intercal$.
The action space, $\tilde{A}$, is the space of the residual action $\tilde{a}$. The transition probability, $\tilde{T}(\tilde{s}_t, \tilde{a}_t, \tilde{s}_{t+1})$, includes both the dynamics of the original environment, i.e. $T(s_t,a_t,s_{t+1})$, and also the dynamics of the base action $\hat{a}_t = \pi_{\mathrm{BeT}}(\cdot | \mathbf{s}_{-K, t})$.

The residual action is predicted by a Gaussian residual policy, $\tilde{a} \sim f_{\textrm{res}}(\tilde{a}|s_t, \hat{a}_t) = \mathcal{N}(\mu, \sigma)$, conditioned on the current state and the base policy's action. The action in the environment, $a_t$, is the sum of the base and residual actions:
\begin{equation}
    a_t = \hat{a}_t + \operatorname{clip}(\tilde{a}_t, -\alpha, \alpha).
\end{equation}
The residual policy is constrained between $[-\alpha, \alpha]$, constraining how much the environment action $a_t$ is allowed to deviate from the base action $\hat{a}$ \cite{rana2023residual,silver2018residual}. For small $\alpha$, the environment action must  be close to the BeT base action, $\hat{a}_t$.

Assuming the action space of $f_{\theta}(\cdot| s_t, \hat{a}_t)$ is restricted to the range $[-\alpha, \alpha]$, we can define the policy $\pi_A(a|\mathbf{s}_{-k, t})$: 
\begin{equation}\label{eq:rpl}
\begin{aligned}
\pi_A &(a|\mathbf{s}_{-k, t}) =&\\ 
  &\tilde{a}_t|_{\tilde{a}_t\sim f_{\mathrm{res}}(\cdot|s_t, \pi_{\mathrm{BeT}}(\cdot|\mathbf{s}_{-k, t}))} + \pi_{\mathrm{BeT}}(\cdot|\mathbf{s}_{-k, t}).
\end{aligned}
\end{equation}
Eq. \ref{eq:rpl} follows the notation of \cite{trumpp2023residual}. Because the BeT policy is non-Markovian, the agent's policy $\pi_A$ is also non-Markovian. However, using the definition of the augmented state, $\tilde{s}$, the input to the AIL residual policy is the current state and the base action predicted by the BeT. Thus, $\pi_A$ is in fact Markovian with respect to the augmented state $\tilde{s}$: 
\begin{equation}\label{eq:markovian_residual}
   \pi_A (a|s_t, \hat{a}_t) = \tilde{a}_t|_{\tilde{a}_t\sim f_{\mathrm{res}}(\cdot|s_t, \hat{a}_t)} + \hat{a}_t.
\end{equation}
Per prior works \cite{silver2018residual, trumpp2023residual}, during online training, the agent's policy during rollouts is given by $\pi_A$ in \eqref{eq:markovian_residual} as the sum of the action from the base policy  $\pi_{\mathrm{BeT}}$ and the action from the residual policy $f_{\mathrm{res}}$~\eqref{eq:rpl}. After pretraining (Section \ref{sec:pretrain}), the  base policy,  $\pi_{\mathrm{BeT}}$, is frozen; then, the agent's policy is improved by updating the residual policy $f_{\mathrm{res}}$ with AIL.

\subsection{Residual Policy Training with AIL}

To train the residual policy, $f_{\mathrm{res}}$, we adapt Adversarial Imitation Learning (AIL) \cite{kostrikov2018discriminator} so that the agent's policy  solves the AIL objective. AIL is defined in terms of a single Markovian policy that is not added to a base policy \cite{ho2016generative}. In this section, we detail our changes to employ AIL to update the residual policy $f_{\mathrm{res}}$.  
 Given the definition of $\pi_A$ \eqref{eq:markovian_residual}, we define the AIL objective for residual policy learning as
\begin{equation} \label{eq:occupancy_residual}
    \underset{f_{\mathrm{res}}}{\operatorname{minimize}} \quad D_{\mathrm{JS}}\left(\rho_{\pi_A}, \rho_{\pi_E}\right)-\lambda H(f_{\mathrm{res}}).
\end{equation}
Similar to the standard AIL objective \cite{kostrikov2018discriminator}, we still aim to minimize the distance between the occupancy measures of the expert's and agent's policies, denoted as $\rho_{\pi_E}$ and $\rho_{\pi_A}$ respectively in Eq.~\eqref{eq:occupancy_residual}. The minimization with respect to $f_{\mathrm{res}}$ is valid since Eq. \eqref{eq:markovian_residual} simply defines a more restrictive class of policies than standard single-policy AIL. Since the contribution from the base policy is deterministic, we regularize the problem  using only the entropy of the residual policy, and the policy update step is replaced with
\begin{equation}\label{eq:sac_obj_residual}
    \max _{f_{\mathrm{res}}} \underset{\tilde{\tau} \sim f_{\mathrm{res}}}{\mathbb{E}}\left[\sum_{t=0}^{\infty} \gamma^t\Bigg(\tilde{R}^E\left(\tilde{s}_t, \tilde{a}_t\right)+\lambda H\left(f_{\mathrm{res}}\left(\cdot \mid \tilde{s}_t\right)\right)\Bigg)\right],
\end{equation}
where $\tilde{\tau}$ represents the trajectory $\tilde{\tau} =(\tilde{s}_0, \tilde{a}_0, ...) $ collected using $f_{\mathrm{res}}$ on $\tilde{\mathcal{M}}$. Thus, $f_{\mathrm{res}}$ is updated using the augmented state and residual action. In residual policy RL \cite{silver2018residual}, the reward is calculated using the action taken in the environment.
Similarly, we  define $\tilde{R}^E$ as AIL's proxy  reward on $\tilde{M}$:
\begin{equation}\label{eq:ail_reward_residual}
\begin{aligned}
       \tilde{R}^E(\tilde{s}_t, \tilde{a}_t)
       &= -\log \left(1-D_\omega^E(s, \hat{a}+\tilde{a})\right),
\end{aligned}
\end{equation}
where $D_\omega^E(s, \hat{a}+\tilde{a})$ is a binary classifier trained to minimize
\begin{equation}\label{eq:disc_loss_residual}
\begin{aligned}
    \mathcal{L}_{ D, \mathcal{D}_E}(\omega)=-&E_{\tau_E\sim \mathcal{D}_E}\left[\log \left(D^E_\omega(s, a)\right)\right]\\&-E_{\tilde{\tau} \sim f_{\mathrm{res}}}\left[\log \left(1-D^E_\omega(s, \hat{a}+\tilde{a})\right)\right],
    \end{aligned}
\end{equation}
which is equivalent to 
\begin{equation}\label{eq:disc_loss_residual2}
\begin{aligned}
    \mathcal{L}_{ D, \mathcal{D}_E}(\omega)=-&E_{\tau_E\sim \mathcal{D}_E}\left[\log \left(D^E_\omega(s, a)\right)\right]\\&-E_{\tau \sim \pi_a}\left[\log \left(1-D^E_\omega(s, a)\right)\right].
    \end{aligned}
\end{equation}
Eq. \eqref{eq:ail_reward_residual} implies that given the pair $(\tilde{s}, \tilde{a})$, the reward is the probability that the state-action pair $(s,a)=(s,\hat{a}+\tilde{a})$ comes from the expert policy, according to the discriminator.  Thus, by iterating between the policy learning objective \eqref{eq:sac_obj_residual} and the discriminator loss \eqref{eq:disc_loss_residual2}, AIL minimizes \eqref{eq:occupancy_residual} to find the $f_{\mathrm{res}}$ that allows $\pi_A$ to match the occupancy measure of the expert $\pi_E$. 

\section{Imitation of Human Racing Gameplay}\label{sec:experiment_setup}
We describe our method to learn from human gameplay in GTS, including the state features, environment, and training.

\subsection{State Feature Extraction and Actions}
The state includes features that were previously  shown as important in RL. We include the following states exactly as described in \cite{fuchs2021super}: 1) linear velocity, $\mathbf{v}_t\in \mathbb{R}^3$, and linear acceleration
$\dot{\mathbf{v}}_t\in \mathbb{R}^3$ with respect to the inertial frame of the vehicle; 2) Euler angle $\theta_t \in (-\pi, \pi]$ between the 2D
vector that defines the agent’s rotation in the horizontal plane
and the unit tangent vector that is tangent to the centerline at
the projection point; 3) a binary flag with $w_t = 1$ indicating
wall contact; and 4) N sampled curvature measurement of the
course centerline in the near future~$\mathbf{c}_t \in \mathbb{R}^N$. 

Additionally, we select features similar to those used in \cite{wurman2022outracing}: 5.) The cosine and sine of the vehicle's current heading, $\cos (\psi)$ and $\sin(\psi)$; and 6.) The relative 2D distance from the vehicle's current position to the left, $\mathbf{e}_{LA, l}$, right, $\mathbf{e}_{LA, r}$, and center, $\mathbf{e}_{LA, c}$, of the track at 5 ``look-ahead points.'' The look-ahead points are placed evenly using a look-ahead time of 2 seconds, i.e., they are spaced evenly over the next 2 seconds of track, assuming the vehicle maintains its current speed. The full state is a vector composed of $s_t=\left[\boldsymbol{v}_t, \dot{\boldsymbol{v}}_t, \theta_t, w_t, \boldsymbol{c}_t,\cos (\psi), \sin(\psi),  \mathbf{e}_{LA, l}, \mathbf{e}_{LA, r}, \mathbf{e}_{LA, c}\right]$.
The state is normalized using the mean and standard deviation of each feature in the demonstrations. 

GTS receives the steering command $\delta\in [-\pi/6, \pi/6]$ rad and a throttle-brake signal $\omega_\tau \in [-1,1]$ where $\omega_\tau=1$ denotes full throttle and $\omega_\tau =-1$ denotes full brake \cite{fuchs2021super,wurman2022outracing}. For all baseline comparisons, the steering command in the demonstrations is scaled to be between $[-1, 1]$.  The agent specifies steering actions between $[-1, 1]$, which are scaled to $\delta\in [-\pi/6, \pi/6]$ before being sent to GTS.

When a residual policy is learned (see BeTAIL and BCAIL in the next section), the residual network predicts $\tilde{a}\in [-1,1]$, and then $\tilde{a}$ is scaled\footnote{The choice to scale the residual policy between [-1,1] is made based on the practice of using scaled action spaces with SAC \cite{haarnoja2018soft}} to $[-\alpha, \alpha]$ and then added to $a=\hat{a}+\tilde{a}$. Since $\hat{a}\in[-1, 1]$, it is possible for $a$ to be outside the bounds of $[-1,1]$, so we clip $a$ before sending the action to GTS. The choice of the hyperparameter $\alpha$ determines the maximum magnitude of the residual action. Depending on the task and the strength of the base policy,  relatively large $\alpha$ allows the residual policy to correct for bad actions \cite{trumpp2023residual}, or small $\alpha$ ensures the policy does not deviate significantly from the base policy \cite{silver2018residual, won2022physics}. In this work, we set $\alpha$ as small as possible so that the policy remains near the offline BeT, which captures non-Markovian human racing behavior. 

\subsection{Environment and Data Collection}
\subsubsection{Gran Turismo Sport Racing Simulator}
We conduct experiments in the high-fidelity PlayStation (PS) game 
Gran Turismo Sport (GTS) ({\footnotesize \url{https://www.gran-turismo.com/us/}}), developed by Polyphony Digital, Inc.  
GTS takes two continuous inputs: throttle/braking and steering. The vehicle positions, velocities, accelerations, and pose are observed. The agent's and demonstrator's state features and control inputs are identical.
 RL achieved super-human performance against human opponents in GTS \cite{wurman2022outracing} but required tuned, dense reward functions to behave ``well'' with  opponents. Rather than  crafting a reward function, we explore if expert demonstrations  can inform a top-performing agent.


 To collect rollouts and evaluation episodes, the GTS simulator runs on a \blue{PS5}, while the agent runs on a separate desktop computer. To accelerate training and evaluation, 20 cars run on the PS, starting at evenly spaced positions on the track \cite{fuchs2021super}. The desktop computer \blue{(Alienware-R13, CPU Intel i9-12900, GPU Nvidia
3090)} communicates with GTS over a dedicated API via an Ethernet as described in \cite{fuchs2021super}.  The API provides the current state of 20 cars and accepts car control commands, which are active until the next command is received. While the GTS state is updated every 60Hz, we limit the control frequency of our agent to 10Hz  \cite{fuchs2021super, wurman2022outracing}.  Unlike prior works that collected rollouts on multiple PSs, we employ a single PS to reduce the desktop's computational burden \blue{and ensure BeTAIL inference meets the 10Hz control frequency}. \blue{During training, rollouts are 50s (500 steps); during evaluation, rollouts are 500s (5,000 steps) to allow sufficient time to complete full laps. Experiments use 8M online environment steps ($\sim$222 hours); with 20 cars collecting data, this results in $\sim$25 hours wall-clock. }

\begin{table}[]
    \centering
    \begin{tabular}{r|lr|l}
    \multicolumn{2}{r}{\textbf{\underline{BeT Pretraining}$\quad$}}  &  \multicolumn{2}{r}{\textbf{\underline{AIL Discriminator Training}}} \\
Layers &4 & Disc. Net. Arch. & [32,32]\\ 
Attention heads & 4 &Updates per iter& 32 \\ 
Embedding dim. & 512 & Learning Rate & 0.005\\
Train context length & 20 & Entropy Scale & 0.001\\
Dropout & 0.1 & Grad. Pen. Scale & 10.0\\
Nonlinearity function & ReLU & Grad. Pen. Target & 1.0\\
Batch size &256 & Demo Batch Size & 2,000 \\
Pretrain Updates & 500,000&Optimizer & Adam\\
 Eval context length & 5&\multicolumn{2}{r}{\textbf{\underline{AIL Policy Training (SAC)}}}\\
Optimizer & Lamb\cite{zheng2022online} & Network Arch. & [256,256] \\ 
Learning rate & 0.0001 & Polyak Update & 0.002\\
Weight decay &  0.0005&Discount Factor & 0.99\\
\multicolumn{2}{r}{}& Gradient Steps & 2500 \\
\multicolumn{2}{r}{}&Replay Buffer & 1M\\
\multicolumn{2}{r}{}&Learning rate & 0.0003 \\
\multicolumn{2}{r}{}&Batch size & 4096\\
\multicolumn{2}{r}{}&Optimizer & Adam\\
\end{tabular}
\begin{tabular}{r|lr|l}
\textbf{Demo Datasets} & \multicolumn{2}{l}{env steps ($\sim$demonstration time)}\\
\hline
BeT(1-track) & 294k(1.6hr)&AIL (Maggiore) & 294k(1.6hr)\\ 
BeT(4-track) & 456k(2.1hr)&AIL (Drag. Tail) & 75k(21min) \\
\multicolumn{2}{r}{}&AIL (Mt. Pan) & 8k(2min)\\ 
    \end{tabular}
    \caption{\blue{Training Hyperparameters and Demonstrations }}
    \label{tab:hyp}
\end{table}

\subsubsection{Training and Baselines}
For each challenge, the BeT, $\pi_{\textrm{BeT}}(\cdot|\mathbf{s}_{-K, t})$, is pretrained on offline human demonstrations from one or more tracks. Then our \textbf{BeTAIL} fine-tunes control by training an additive residual policy, $f_{\mathrm{res}}$,  with AIL on the downstream environment and human demonstrations.  The residual policy is constricted between $[-\alpha,\alpha]$, where $\alpha$ is specified individually for each challenge. Training hyperparameters are listed in Table \ref{tab:hyp}. In Fig. \ref{fig:laptime}a-c,  \textcolor{myred}{\textbf{red}} corresponds to demonstrations for the BeT and \textcolor{mygreen}{\textbf{green}} corresponds to the downstream environment and demonstrations for the residual AIL policy.
The \textbf{BeT} baseline uses the BeT policy, $\pi_{\textrm{BeT}}(\cdot|\mathbf{s}_{-K, t})$, with the pretraining demonstrations \textcolor{myred}{\textbf{red}}. \textbf{AIL} consists of a single policy, $\pi_{A}(a|s)$, trained via AIL using only the downstream demonstrations and environment in \textcolor{mygreen}{\textbf{green}}. 
\textbf{BCAIL} trains a BC policy, $\pi_{A}(a|s)$, on the offline demonstrations (in \textcolor{myred}{\textbf{red}}) and then fine-tunes a residual policy in the downstream environment (in \textcolor{mygreen}{\textbf{green}}) following the same training scheme as BeTAIL. \blue{Table \ref{tab:hyp} lists hyperparameters for BeT pretraining and AIL training.}

Agents use soft actor-critic (SAC) \cite{haarnoja2018soft} for the reinforcement learning algorithm in adversarial learning (i.e. for AIL, BCAIL, and BeTAIL).
Since SAC is off-policy, the replay buffer can include historical rollouts from many prior iterations. However, the discriminator, and thus the reward associated with a state-action pair, may have changed. When sampling data from the replay buffer for policy and Q-network training, we re-calculate the reward for each state-action pair with the current discriminator network. Finally, discriminator overfitting can deteriorate AIL performance, so for all baselines, AIL employs two discriminator regulators: gradient penalty and entropy regularization \cite{orsini2021matters}.

\subsubsection{Demonstrations}

All demonstrations were recorded from \textit{different}, \textit{expert-level} human players GTS participating in real gameplay, ``time-trial'' competitions with the Audi TT Cup vehicle. 
Each demonstration contains a full trajectory of states and actions, around a single lap of the track. The trajectories were recorded at a frequency of 60Hz, which we downsample to 10Hz (the agent's control frequency). Each trajectory (i.e. lap) contains approximately 7000 timesteps, \blue{which are split into 500-step segments to match the length of the training episodes.} Listed in Fig. \ref{fig:laptime}a-c \blue{and Table \ref{tab:hyp}}, we test three training schemes, either finetuning the BeT on the same race track as pretraining or on new, unseen tracks. 



\begin{figure}[b]
    \centering
    \vspace{6pt}
        \includegraphics[width = .95\columnwidth]{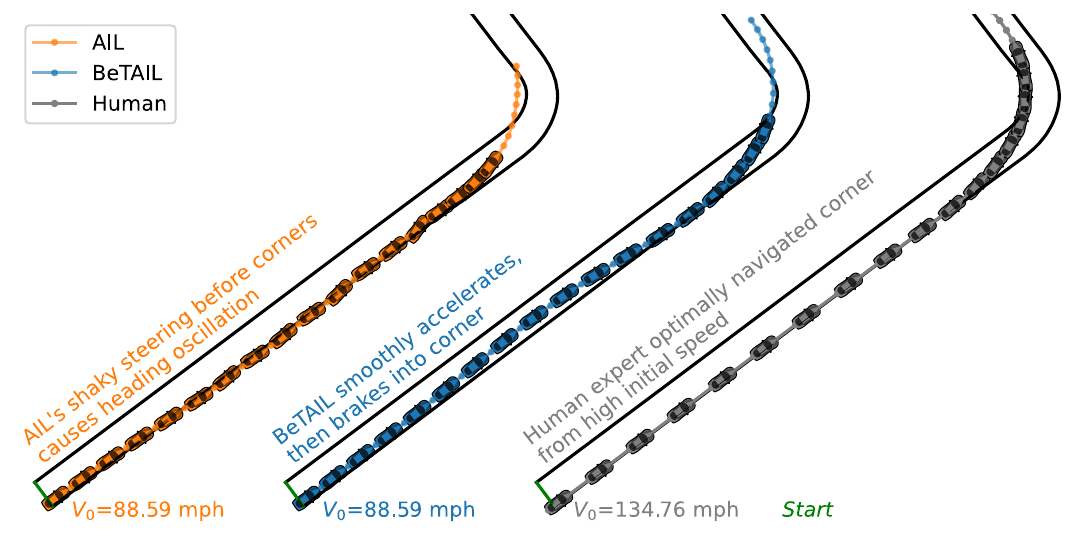}
    \caption{Agent trajectories on Lago Maggiore. We deliberately set AIL and BeTAIL to start at a lower initial speed than the human. Car drawing is placed at the vehicle's location and heading every 0.4s. See  website for the animated version.}
    \label{fig:maggiore_trajectories}
\end{figure}

\section{Racing Experiment Results}\label{sec:results}
There are three challenges with distinct pretraining datasets and downstream environments. For evaluation, 20 cars are randomly placed evenly on the track on one PS. Each car's initial speed is set to the expert's speed at the nearest position in a demonstration. Each car has 500 seconds (5000 steps) to complete a full lap. In Fig. \ref{fig:laptime}d-f, we provide two evaluation metrics during training: the proportion of cars that finish a lap (top) and the average lap times of cars that finish (bottom). Higher success rates and lower lap times indicate better performance.  Error bars show the total standard deviation across the 20 cars and 3 seeds. Fig. \ref{fig:laptime}g-i provide the mean$\pm$standard deviation of the best policy's lap time and absolute change in steering $|\delta_t-\delta_{t-1}|$, as RL policies can exhibit undesirable shaky steering behavior \cite{wurman2022outracing}. Videos of trajectories and GTS game are provided at {\footnotesize \url{https://sites.google.com/berkeley.edu/BeTAIL/home}}.

\subsection{\textbf{Lago Maggiore Challenge}: Fine-tuning in the same environment with BeTAIL} 
\begin{figure*}
    \centering
        \subfloat[Lago Maggiore Challenge: Training Scheme]{\vtop{\vskip0pt\hbox{
    \includegraphics[width = .3\textwidth]{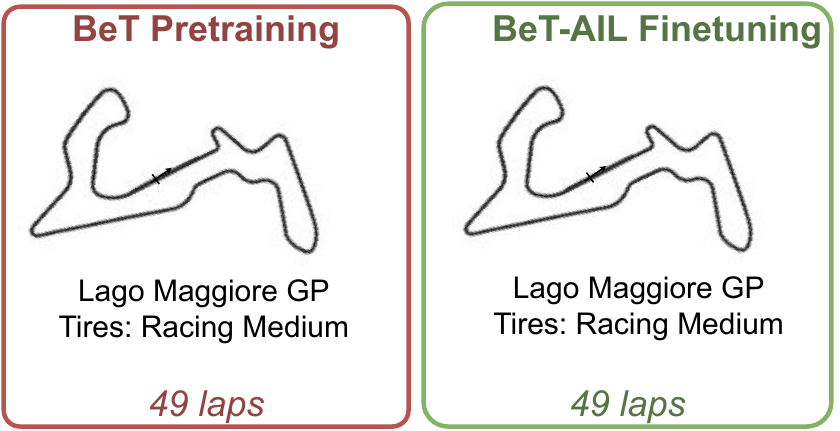}}}
    }
    \subfloat[Dragon Tail Challenge: Training Scheme]{\vtop{\vskip0pt\hbox{
    \includegraphics[width = .3\textwidth]{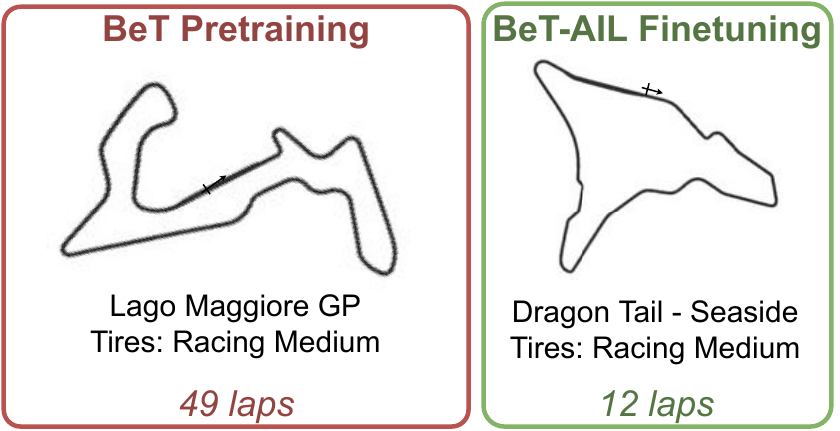} }
    }}
    \subfloat[Mount Panorama Challenge: Training Scheme]{\vtop{\vskip0pt\hbox{
    \includegraphics[trim=0pt 28pt 0pt 0pt, clip, width = .3\textwidth]{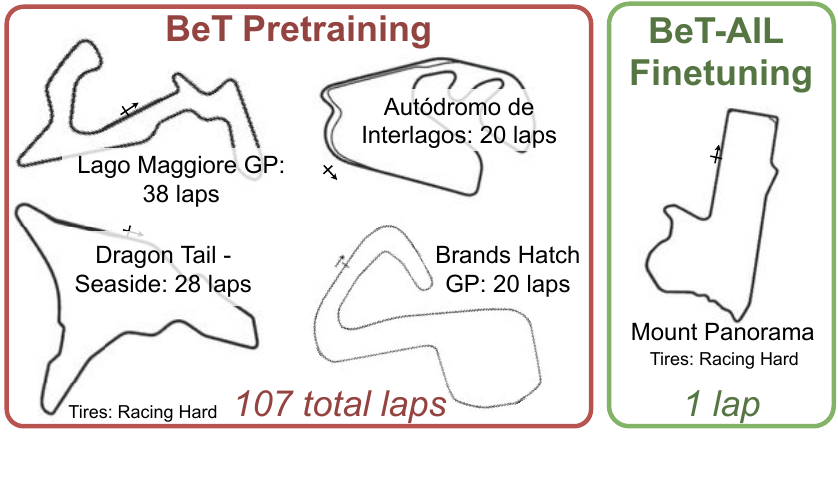}}}
    }\\
    \subfloat[Training on Lago Maggiore (mean $\pm$ std)]{\vtop{\vskip0pt\hbox{
    \includegraphics[width = .3\textwidth]{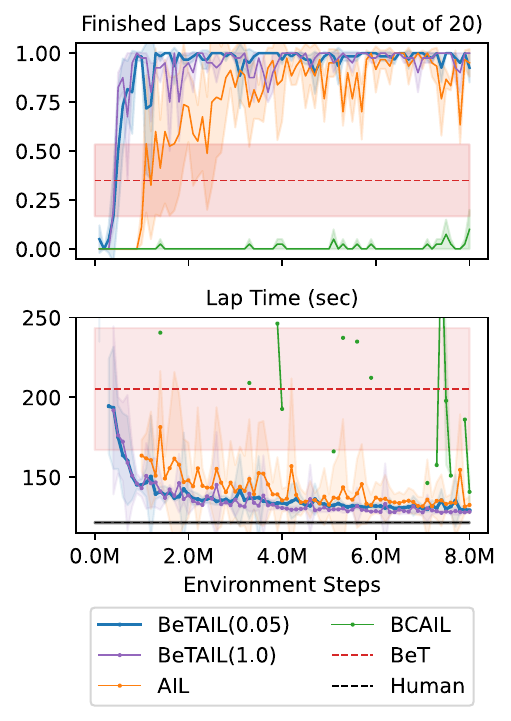}}}
    }
    \subfloat[Training on Dragon Tail (mean $\pm$ std)]{\vtop{\vskip0pt\hbox{
        \includegraphics [width = .3\textwidth]{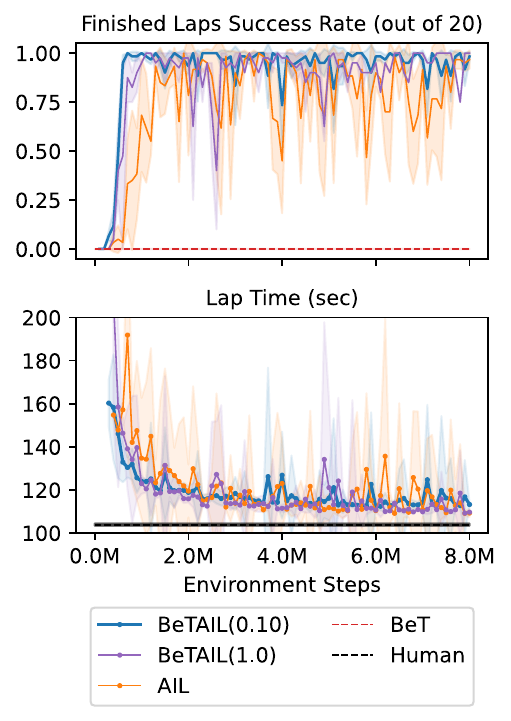} }
    }}
    \subfloat[Training on Mount Panorama (mean $\pm$ std)]{\vtop{\vskip0pt\hbox{
        \includegraphics[width = .3\textwidth]{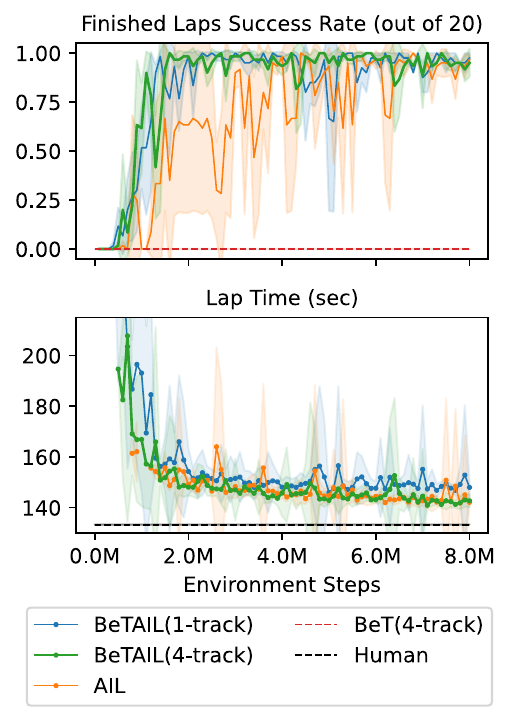}}}
    }\\
    \subfloat[Results on Lago Maggiore Track]{\makebox[.33\textwidth]{
    \centering
            \scriptsize
            \begin{tabular}{|c|c|c|}
                \hline
                 & \textbf{Lap } & \textbf{Steering Change} \\ \textbf{Algorithm} & \textbf{Time} (s) & $|\delta_t - \delta_{t-1}|$ (rad)\\
                \hline  
BeTAIL(0.05)  &  129.2$\pm$0.7 & 0.014$\pm$0.011 \\
             BeTAIL(1.0)  &  \textbf{127.7$\pm$0.7} & 0.068$\pm$0.085 \\
AIL  &  131.6$\pm$4.1 & 0.081$\pm$0.091 \\
BCAIL  &  140.7$\pm$8.6 & 0.022$\pm$0.021 \\
BeT  &  205.2 $\pm$38 & \textbf{ 0.0058$\pm$0.010} \\
\hline 
\textit{Human}  &  \textit{121.5 $\pm$0.4} & \textit{0.0067$\pm$0.011} \\
                \hline
            \end{tabular}}}
            \subfloat[Results on Dragon Tail Track]{\makebox[.33\textwidth]{
    \centering
            \scriptsize
            \begin{tabular}{|c|c|c|}
                \hline
                 & \textbf{Lap } & \textbf{Steering Change} \\ \textbf{Algorithm} & \textbf{Time} (s) & $|\delta_t - \delta_{t-1}|$ (rad)\\
                \hline  
BeTAIL(0.10)  &  112.1$\pm$0.8 & \textbf{0.024$\pm$0.019} \\
BeTAIL(1.0)  &  109.2$\pm$0.3 & 0.083$\pm$0.094 \\
AIL  &  \textbf{109.1$\pm$0.7} & 0.077$\pm$0.075 \\
BeT  &  unfinished & 0.011$\pm$0.019 \\
\hline 
\textit{Human}  &  \textit{103.9 $\pm$0.5} & \textit{0.0018$\pm$0.0061} \\
                \hline
            \end{tabular}}}
                        \subfloat[Best Results on Mount Panorama Track ($\alpha$=0.2)]{\makebox[.33\textwidth]{
    \centering
            \scriptsize
            \begin{tabular}{|c|c|c|}
                \hline
                 & \textbf{Lap } & \textbf{Steering Change} \\ \textbf{Algorithm} & \textbf{Time} (s) & $|\delta_t - \delta_{t-1}|$ (rad)\\
                \hline  
BeTAIL(1-trk)  &  146.0$\pm$1.9 & 0.044$\pm$0.036 \\
BeTAIL(4-trk)  &  \textbf{140.9$\pm$1.2} & \textbf{0.039$\pm$0.033} \\
AIL  &  141.8$\pm$1.4 & 0.087$\pm$0.088 \\
BeT(4-trk) &  nan $\pm$0.0 & 0.014$\pm$0.024 \\
\hline 
\textit{Human}  &  \textit{133.3 $\pm$0.0} & \textit{0.0067$\pm$0.013} \\
                \hline
            \end{tabular}}}
    \caption{Experimental results on three racing challenges. (a) Lago Maggiore challenges pretrains the BeT on the same demonstrations and downstream environments. (b) Dragon Tail transfers the BeT policy to a new track with BeTAIL finetuning. (c) The Mount Panorama challenge pretrains the BeT on a library of 4 tracks, and BeTAIL finetunes on an unseen track. (d)-(f) evaluation of mean (std) success rate to finish laps and mean (std) of lap times. (g)-(i) Best policy's mean $\pm$ std lap time and change in steering from previous time step. \blue{(8M steps $\approx$ 25 hours w/ 20 cars collecting data)}
    }
    \label{fig:laptime}
\end{figure*}

We test if BeTAIL can fine-tune the BeT policy  on the same track as the downstream environment (Fig. \ref{fig:laptime}a). BeTAIL(0.05) employs a small residual policy, $\alpha = 0.05$, \blue{so that the agent's action is close to the action predicted by the BeT ~\cite{rana2023residual}}. The ablation BeTAIL(1.0) uses a large residual policy, $\alpha = 1.0$. There are 49 demonstrations from different human players on the Lago Maggiore track. The BeT is trained on the 49 trajectories, then we run BeTAIL on the same 49 trajectories and the same environment as the demonstrations. BCAIL follows the same training scheme  ($\alpha=0.05$). AIL is trained on the same 49 trajectories. 

Fig. \ref{fig:laptime}d evaluates each agent during training. BeTAIL outperforms all other methods and rapidly learns a policy that can consistently finish a full lap and achieve the lowest lap time. AIL  eventually learns a policy that navigates the track and achieves a low lap time; however, even towards the end of the training, AIL is less consistent, and multiple cars may fail to finish a full lap (higher standard deviation in the top of Fig. \ref{fig:laptime}d). The other baselines perform poorly on this task, so they are not tested on the other, more difficult challenges. A residual BC policy (BCAIL) worsens performance, since the BC policy performs poorly in the online environment. Impressively, the BeT finishes some laps even though it is trained exclusively on offline data; but, the lap time is poor compared to BeTAIL and AIL, which exploit  online rollouts.

In Fig. \ref{fig:laptime}g, BeTAIL(1.0) achieves a lower lap time than BeTAIL(0.05), since larger $\alpha$ allows the residual policy to apply more correction to the BeT base policy. However, BeTAIL(1.0) is prone to oscillate steering back and forth, as indicated by the higher deviation in the steering command between timesteps. We hypothesize that it is because the non-Markovian human-like behavior diminishes as $\alpha$ becomes larger. It is supported by the observation that the Markovian AIL policy also tends to have extreme changes in the steering command. Fig. \ref{fig:maggiore_trajectories} provides insight AIL's extreme steering rates. We deliberately initialize the race cars at a lower speed than the human to test the robustness of AIL and BeTAIL agents. The heading of the AIL agent oscillates due to its tendency to steer back and forth, which causes the AIL agent to lose control and collide with the corner. Conversely, BeTAIL smoothly accelerates then brakes into the corner.

\subsection{\textbf{Dragon Tail Challenge:} Transferring the BeT policy to another track with BeTAIL}
The second challenge tests if BeTAIL can fine-tune the BeT  when the downstream environment is different from the BeT demonstrations (Fig. \ref{fig:laptime}b). The BeT from the Lago Maggiore Challenge is used as the base policy; however, the downstream environment is a different track (Dragon Trail), which has 12 demonstration laps for AIL training.   The residual policy is allowed to be larger, BeTAIL(0.10), since the downstream environment is different than the BeT pretraining  dataset. The vehicle dynamics are unchanged. 

The results for training and evaluation on the Dragon Tail track are given in Fig. \ref{fig:laptime}e/h. Again, BeTAIL  employs the BeT to guide policy learning and quickly learns to navigate the track at a high speed. Additionally, small $\alpha$  ensures that non-Markovian human behavior is preserved, resulting in smooth steering. Conversely, AIL  learn policies that are capable of achieving low lap times; however, they exhibit undesirable rapid changes in steering and are significantly more prone to fail to finish a lap (top in Fig. \ref{fig:laptime}e). The pretrained BeT, which was trained from demonstrations on a different track, is unable to complete any laps.
\begin{figure*}
    \centering
    \subfloat[Training on Lago Maggiore Track]{\includegraphics[width=.65\textwidth,valign=m]{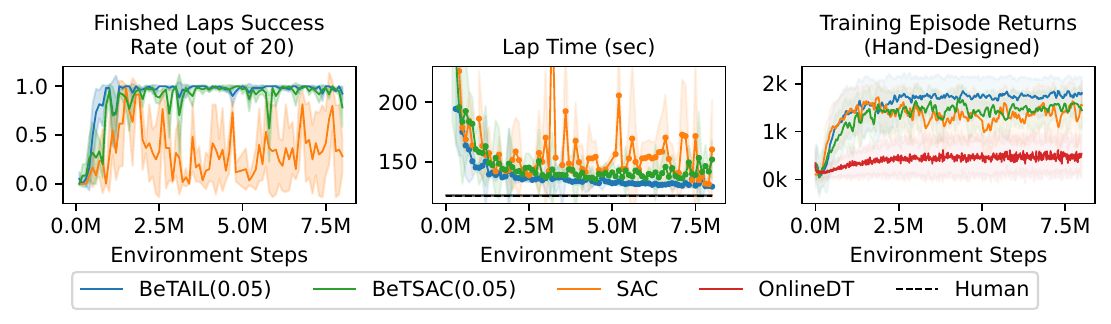}}
    \hfill
    \subfloat[Policy performance after 8M environment steps. \\ \scriptsize $\dagger$: Results after a total of 17M steps (1 random seed)\\$\ddagger$: Overal finished lap success rate in final 1M training steps
    ]{{\makebox[.33\textwidth][r]{
    \centering
            \scriptsize
            \setlength\tabcolsep{4pt}
            \begin{tabular}{|c|c|c|c|}
                \hline
                 &\textbf{Success}& \textbf{Lap } & \textbf{Steering} \\ \textbf{Algorithm} & \textbf{Rate}$^\ddagger$&\textbf{Time} (s) & \textbf{Change}(rad)\\
                \hline  
BeTAIL(0.05)& \textbf{100}\% &  129.2$\pm$0.7 & 0.014$\pm$0.011 \\
BeTSAC(0.05) & 94.2\% &  135.3$\pm$1.0 & 0.013$\pm$0.011 \\
SAC &41.5\% &  131.2$\pm$7.1 & 0.093$\pm$0.118 \\
SAC$^\dagger$ & 95\% &  \textbf{126.1$\pm$0.2} & 0.067$\pm$0.066 \\
BeTAIL(0.05)$^\dagger$ &\textbf{100\%}&  \textbf{125.9$\pm$1.3} & 0.014$\pm$0.011 \\
\hline 
\textit{Human} & -&  \textit{121.5 $\pm$0.4} & \textit{.0067$\pm$.011} \\
                \hline
            \end{tabular}}}
}
    \caption{\blue{Ablation study on Lago Maggiore (Fig. \ref{fig:laptime}a). \textbf{SAC} trains a Markov policy, replacing the AIL reward with the reward in \eqref{eq:rl_reward}. \textbf{BeTSAC(0.05)} replaces the AIL residual policy finetuning step with SAC finetuning using \eqref{eq:rl_reward}. 3 seeds unless noted.}}
    \label{fig:rl_ablation}
\end{figure*}

\subsection{\textbf{Mount Panorama Challenge:} Learning a multi-track BeT policy and solving an unseen track with BeTAIL}
Finally, a BeT policy is trained on a library of trajectories on four different tracks: Lago Maggiore GP (38 laps), Autodromo de Interlagos (20 laps), Dragon Tail - Seaside (28 laps), and Brands Hatch GP (20 laps). For BeTAIL fine-tuning, BeTAIL is trained on a single demonstration on the Mount Panorama Circuit. Due to the availability of trajectories, there is a slight change in vehicle dynamics from the first two challenges due to the use of different tires (Racing Hard); the vehicle dynamics in downstream training and evaluation employ the same Racing Hard tires as the library of trajectories. The Mount Panorama circuit has more complex course geometry with hills and sharp banked turns than the pretraining tracks; thus, the residual policy is larger to correct for errors in the offline BeT.
In Fig. \ref{fig:laptime}f,\textbf{ BeTAIL(4-track)} indicates the BeT is trained on the library of trajectories (Fig. \ref{fig:laptime}c) with $\alpha=0.2$. As an ablation, we compare \textbf{BeTAIL(1-track)} with $\alpha=0.2$, where the pretrained BeT is the one used in Fig. \ref{fig:laptime}a/b with Racing Medium tires, which have a lower coefficient of friction than the Racing Hard tires in the downstream environment. 

 As with previous challenges, both BeTAIL(1-track) and BeTAIL(4-track)  navigate the track with less environment interactions than AIL alone. BeTAIL(4-track) achieves faster lap times and smoother steering than AIL, indicating that BeT pre-training on a library of tracks can assist learning on new tracks. 
BeTAIL(1-track) exhibits a slight performance drop when transferring between different vehicle dynamics compared to BeTAIL(4-track). However, BeTAIL(1-track) still accelerates training to achieve a high success rate faster than AIL alone. In our website's videos, all agents struggle at the beginning of the track, but BeTAIL and AIL  navigate the track rapidly despite the complicated track geometry and hills. AIL exhibits undesirable shaking behavior, but BeTAIL smoothly navigates  with the highest speed.


\subsection{\blue{\textbf{Reinforcement Learning Ablation} on Lago Maggiore}}
\blue{BeTAIL is a reward-free IL algorithm. In this subsection, we further compare it against several RL baselines that have access to explicit environment rewards. 
We employ a reward identical to \cite{fuchs2021super}: 
\begin{equation}\label{eq:rl_reward}
    R(s_t, a_t)\doteq (cp_t - cp_{t-1}) - c_w \|\mathbf{v}_t\|^2 \mathbb{I}_t^{o.c.}, 
\end{equation}
where the first term, $(cp_t - cp_{t-1})$, represents the course progress along the track centerline, and the second term is an penalty given by the indicator flag $\mathbb{I}_t^{o.c.}$ when the vehicle goes off course (outside track boundaries). The weight $c_w$ is set identical to \cite{fuchs2021super}. \textbf{SAC} is trained on \eqref{eq:rl_reward} with soft actor critic \cite{haarnoja2018soft} with identical network architecture  as the AIL baseline. The ablation \textbf{BeTSAC} is identical to BeTAIL except the residual policy is trained with SAC on \eqref{eq:rl_reward} instead of AIL.}

\blue{The Online Decision Transformer (\textbf{OnlineDT}) \cite{zheng2022online},  includes a return-to-go (rtg) conditioning in addition to the state-action inputs to the BeT. Since the DT includes rtg, collecting online rollouts improves performance \cite{zheng2022online}. Our OnlineDT baseline is pretrained with the same data and number of iterations as the BeT, and then online rollouts are collected and labeled with \eqref{eq:rl_reward}. Following \cite{zheng2022online}, the rollout rtg is 5000, which is double the maximum expected rewards for the 500-step training rollouts and 500-step offline dataset trajectory segments. Prior work does not examine rtg with evaluation episodes that are longer than training rollouts \cite{zheng2022online}. Thus, we only evaluate OnlineDT in the 500-step training episodes and do not test 5000-step lap time evaluations. }

\blue{Fig. \ref{fig:rl_ablation} shows that BeTAIL and BeTSAC reduce the number of environment steps required for training in comparison to SAC. BeTSAC performs worse than BeTAIL, and SAC actually performs worse than AIL (see Fig. \ref{fig:laptime}d) within 8M environment steps; we hypothesize that it is because the AIL objective provides better guidance than the heuristically shaped dense rewards. OnlineDT struggles to maximize rewards even in training rollouts, likely because rtg conditioning is insufficient for stochastic environments where rewards are not guaranteed~\cite{gao2023act}. The results call for further exploration of best practices for online training of Transformer architectures~\cite{li2023survey}.}

\blue{
Fig. \ref{fig:rl_ablation}b provides a summary of the results at the end of training. Longer training (17M environment steps/$\sim$2.5 days wall-clock)
improves the SAC lap time near that of BeTAIL. However, SAC is less stable, as shown by an average of the lap success rate in the final 1M steps of training. BeTAIL(0.05) finishes every evaluation lap after only 8M training steps, whereas SAC still fails to always finish every lap after 17M training steps.   Previous SAC successes required custom rewards, custom training regimes and starting positions, and extensive environment interactions lasting multiple days for training on multiple devices \cite{wurman2022outracing, fuchs2021super}.  Conversely, BeTAIL exploits the pretrained BeT (using only a couple hours of offline data) to  accelerate training and learn high-performing policies from demonstrations.}

\section{Conclusion}\label{sec:conclude}
We proposed BeTAIL, a Behavior Transformer augmented with residual Adversarial Imitation Learning. BeTAIL leverages sequence modeling and online imitation learning to learn racing policies from human demonstrations. In three experiments in the high-fidelity racing simulator Gran Turismo Sport, we show that BeTAIL can leverage both demonstrations on a single track and a library of tracks to accelerate downstream AIL on unseen tracks. BeTAIL policies exhibit smoother steering and reliably complete racing laps. In a small ablation, we show that BeTAIL can accelerate learning under minor dynamics shifts when the BeT is trained with different tires and tracks than the downstream residual AIL.

\textit{Limitations and Future Work:} BeTAIL employs separate pre-trained BeT and residual AIL policy networks. The residual policy and the discriminator network are Markovian, rather than exploiting the sequence modeling in the BeT. Future work could explore alternate theoretical frameworks that improve the BeT action predictions themselves. Also,  sequence modeling could be introduced into AIL frameworks to match the policy's and experts' trajectory sequences instead of single-step state-action occupancy measures. Finally, there is still a small gap between BeTAIL's lap times and the lap times achieved in the expert demonstrations.  Future work will explore how to narrow this gap with improved formulations or longer training regiments.
%


\bibliographystyle{IEEEtran} 
\bibliography{IEEEabrv,mybibfile}

\end{document}